\documentclass[letterpaper]{article} 
\usepackage{aaai23}  
\usepackage{times}  
\usepackage{helvet}  
\usepackage{courier}  
\usepackage[hyphens]{url}  
\usepackage{graphicx} 
\usepackage{natbib}  
\usepackage{caption} 
\usepackage{algorithm}
\usepackage{algorithmic}

\usepackage{newfloat}
\usepackage{listings}
\usepackage{bbding}
\usepackage{bm}
\usepackage{amsmath}
\usepackage{makecell}
\usepackage{natbib}
\setcitestyle{authoryear,round}
\usepackage{newfloat}
\usepackage{listings}
\usepackage{amssymb}
\usepackage{hyperref}
\usepackage[capitalize]{cleveref}

\DeclareCaptionStyle{ruled}{labelfont=normalfont,labelsep=colon,strut=off} 
\floatstyle{ruled}
\newfloat{listing}{tb}{lst}{}
\floatname{listing}{Listing}
\title{GAM : Gradient Attention Module of Optimization for Point Clouds Analysis}
\author{
    Haotian Hu\textsuperscript{\rm 1}, Fanyi Wang\textsuperscript{\rm 2}\thanks{Corresponding author}, Jingwen Su\textsuperscript{\rm 2}, Hongtao Zhou\textsuperscript{\rm 1}, Yaonong Wang\textsuperscript{\rm 1}, Laifeng Hu\textsuperscript{\rm 1}, Yanhao Zhang\textsuperscript{\rm 2}, Zhiwang Zhang\textsuperscript{\rm 3}\thanks{Corresponding author}
}
\affiliations{
    \textsuperscript{\rm 1} Zhejiang Leapmotor Technology CO., LTD.\\
    \textsuperscript{\rm 2} OPPO Research Institute\\
    \textsuperscript{\rm 3} The University of Sydney\\
    
    \{hu\_haotian, zhou\_hongtao, wang\_yaohong, hu\_laifeng\}@leapmotor.com, \{wangfanyi,sujingwen,zhangyanhao\}@oppo.com,\\  zhiwang.zhang@sydney.edu.au 
}

\usepackage{bibentry}

\begin{document}

\maketitle

\begin{abstract}
In point cloud analysis tasks, the existing local feature aggregation descriptors (LFAD) are unable to fully utilize information in the neighborhood of central points.  Previous methods rely solely on Euclidean distance to constrain the local aggregation process, which can be easily affected by abnormal points and cannot adequately fit with the original geometry of the point cloud.  We believe that fine-grained geometric information (FGGI) is significant for the aggregation of local features.  Therefore, we propose a gradient-based local attention module, termed as Gradient Attention Module (GAM), to address the aforementioned problem.  Our proposed GAM simplifies the process that extracts gradient information in the neighborhood and uses the Zenith Angle matrix and Azimuth Angle matrix as explicit representation, which accelerates the module by 35X.  Comprehensive experiments were conducted on five benchmark datasets to demonstrate the effectiveness and generalization capability of the proposed GAM for 3D point cloud analysis.  Especially on S3DIS dataset~\citep{armeni20163d}, GAM achieves the best performance among current point-based models with mIoU/OA/mAcc of 74.4\%/90.6\%/83.2\%, respectively. Code to reproduce our results is available at \url{https://github.com/hht1996ok/GAM }.

\end{abstract}

\section{Introduction}
In recent years, point cloud analysis has become a hot topic in academia and industry due to the rapid development of autonomous driving and indoor robotics. Considering that point cloud is unordered, sparse, and irregular, traditional methods for 2D image processing cannot be directly applied to point clouds. PointNet~\citep{qi2017pointnet} is a pioneering work that uses Multi-Layer Perceptron (MLP) to learn point features independently. Qi et al. proposed PointNet++~\citep{qi2017pointnet++}, which introduces local features to point cloud analysis models for further performance improvement of point cloud analysis models. Recent works present some promising results, using convolutional layers \citep{boulch2020convpoint, thomas2019kpconv, xiang2021walk}, graph structures~\citep{wang2019dynamic, xu2020grid}, MLP~\citep{ma2022rethinking}, or attention mechanisms~\citep{guo2021pct, zhao2021point} for point cloud analysis. Among them, local feature aggregation descriptors (LFAD) play an important role. 

\begin{figure}[t]
    \centering
    \includegraphics[width=\linewidth]{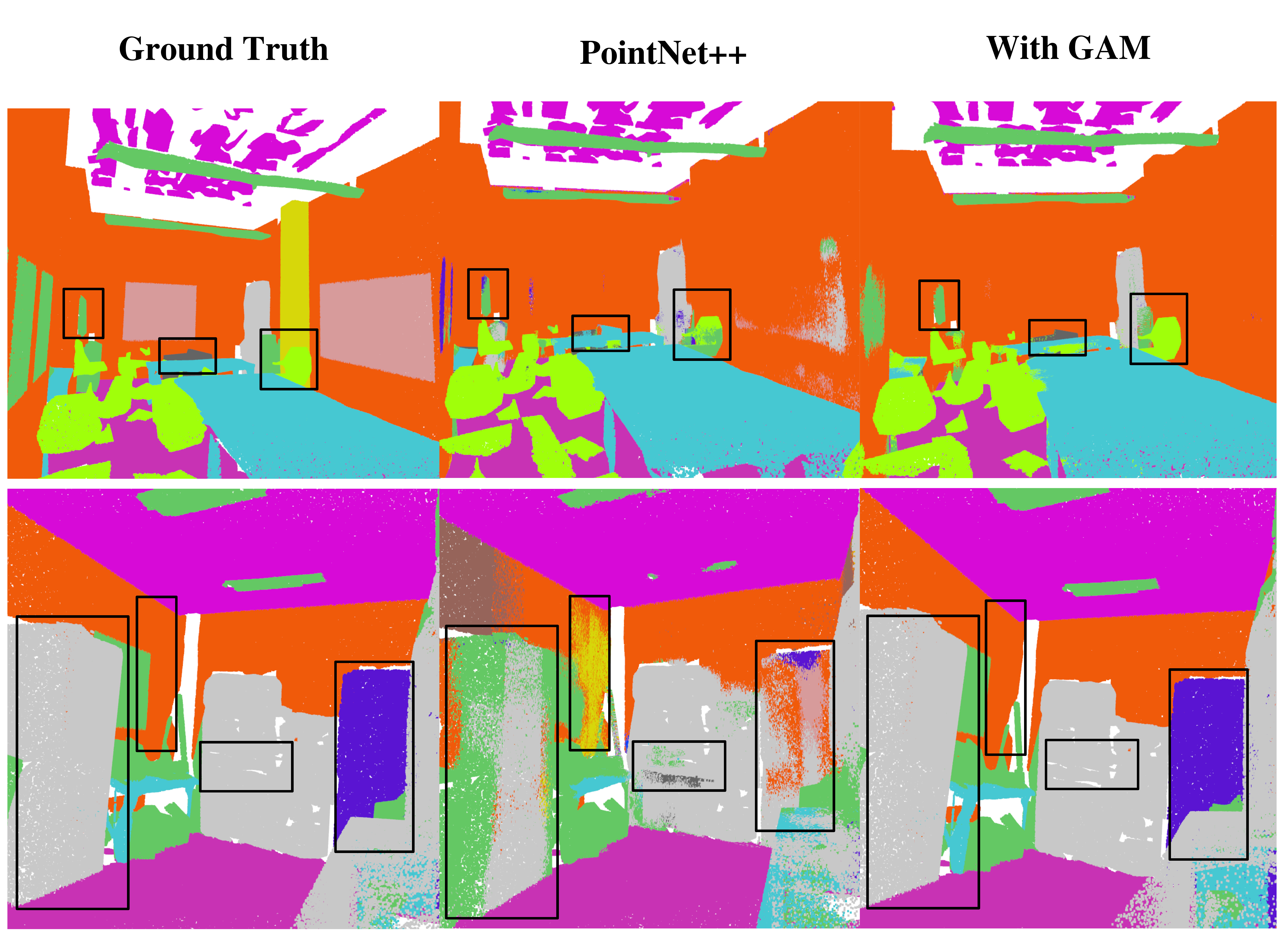}
    \caption{S3DIS benchmark visualization results, from left to right are ground truth, PointNet++~\citep{qi2017pointnet++} and the results after adding the gradient attention module (GAM).}
    \label{fig: fig1}
\end{figure}

However, existing LFAD cannot effectively distinguish points in a point cloud neighborhood, and thus are unable to learn finer semantic information of the point cloud. This observation motivates us to consider the attention mechanism within a point cloud neighborhood. The previous works treat all points in the neighborhood as equally important~\citep{wang2019dynamic}, or only use distance information to constrain aggregation process~\citep{lan2019modeling, thomas2019kpconv, ma2022rethinking}, ignoring deeper geometric relationships within the neighborhood. These operations include too much outlier information in the local feature aggregation process and impede the model to conform the original geometry of the point cloud. Therefore, we propose a novel gradient attention module (GAM) that utilizes neighboring gradient information to better constrain the aggregation process of the neighborhood features. As shown in Figure~\ref{fig: fig1}, with gradient information, our proposed method is enabled to predict clearer object boundaries.

\begin{figure}[t]
    \centering
    \includegraphics[width=\linewidth]{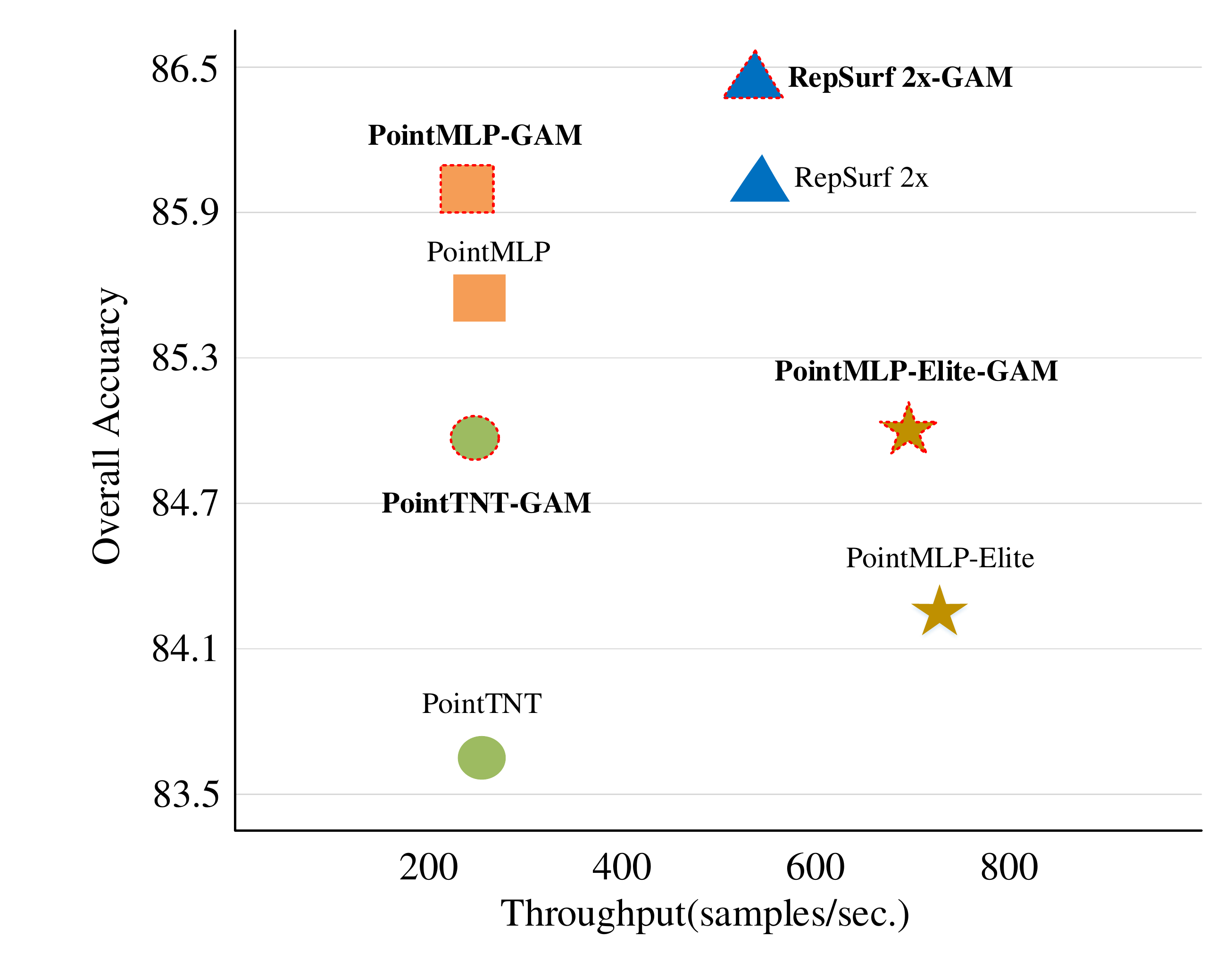}
    \caption{Overall Accuracy (OA) and throughput comparison plots of 3D shape classification experiment results on  PointMLP~\citep{ma2022rethinking}, PointMLP-Elite~\citep{ma2022rethinking}, PointTNT~\citep{berg2022points} and RepSurf 2x~\citep{ran2022surface} models before and after adding GAM in ScanObjectNN~\citep{uy2019revisiting}.}
    \label{fig: fig2}
\end{figure}

In addition, we find that the gradient calculation method~\citep{pauly2003point} based on the local surface fitting method is very slow, and hinders real-time inference after adding gradient information. To solve this problem, we propose to simplify the calculation of gradient information in the neighborhood, by converting gradient information to an explicit representation of the Zenith Angle and Azimuth Angle between the center point and its neighboring points. Our proposed method can accelerate computation speed by 35 times. As shown in Figure~\ref{fig: fig2}, GAM is a plug-and-play module, which effectively improves the performance of baseline methods while maintaining a similar inference speed.

\begin{figure*}[t]
    \centering
    \includegraphics[width=\linewidth]{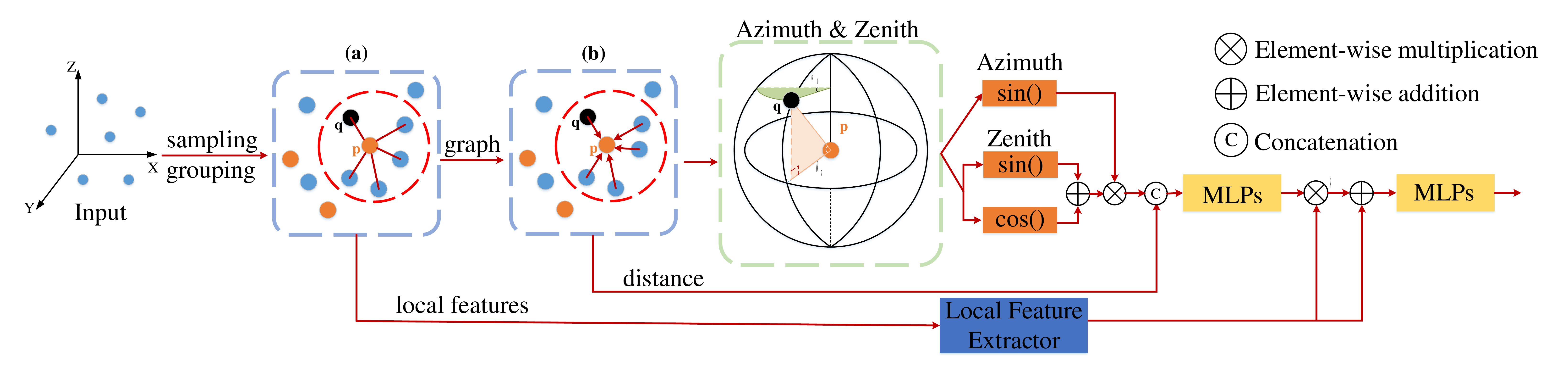}
    \caption{Schematic diagram of GAM structure, where $p$ represents one center point, $q$ represents a particular neighbor point of $p$, and LFE denotes the local feature extractor of different baselines. Module inputs are original positional information of the point cloud and features. Step (a) searches the center points of the point cloud and their corresponding neighborhood points; step (b) builds a directed graph to obtain the relative position vector between the center point and its neighborhood points. Then GAM calculates the zenith and azimuth angles in the neighborhood and uses them to construct a gradient attention matrix.}
    \label{fig: fig3}
\end{figure*}

Our proposed GAM is portable and can be added to previous state-of-the-art methods with a few lines of code. We conduct experiments on 3D semantic segmentation task using S3DIS dataset~\citep{armeni20163d}, 3D shape classification task using ScanObjectNN dataset~\citep{uy2019revisiting} and ModelNet40 dataset~\citep{wu20153d}, 3D part segmentation task using ShapeNet~\citep{yi2016scalable}, and 3D object detection task on KITTI dataset~\citep{geiger2013vision}. Experiment results demonstrate that GAM is an effective module and applicable to a wide range of point cloud analysis tasks with good performance improvements for various models. 

The contributions of this paper are summarised as follows:


\begin{itemize}
\item We propose a lightweight and efficient gradient attentive module (GAM). To the best of our knowledge, gradient information is the first time to be introduced into the vector of locally aggregated descriptors of point cloud neighborhood features.

\item The relationship between gradient information and zenith angle and azimuth angle in the point cloud neighborhood is established through mathematical representation. The gradient calculation process is simplified to the calculation of the zenith angle and azimuth angle of neighborhood points. Thus, the computation speed of GAM is effectively improved.

\item Comprehensive experiments on five benchmarks demonstrate that our proposed gradient attention module can effectively boost performances of state-of-the-art methods within limited additional memory consumption and inference time. In addition, our proposed GAM can be used in various 3D tasks such as 3D semantic segmentation, 3D shape classification, 3d object detection, and 3D part segmentation.

\end{itemize}


\section{Related Work}\label{Sec:Literature}

The point cloud analysis task starts with learning the embedding of each point, then extracts global embedding from the whole point cloud using local aggregation methods, and finally feeds global embedding to branches of each task. Due to the disorderly nature of point clouds, some previous works attempt to project point clouds into regular voxels~\citep{zhou2018voxelnet, wu20153d, yan2018second} or multiple views~\citep{su2015multi, wei2020view, liang2018deep}. These methods significantly improve the computational speed, but lose information during the projection process and undermine model accuracy severely. In contrast, point-based methods directly use original point cloud information as input and employ various well-designed local feature aggregation descriptors~\citep{thomas2019kpconv, lan2019modeling, yang2019modeling, komarichev2019cnn}. Meanwhile, some previous works~\citep{wang2019graph, chen2021GAPointNet, wang2021PointAttN, cui2021Geometric, wu2022casa} use the attention mechanism to extract the feature of the point cloud.

\subsection{Multi-view and Voxel-based Approaches}
Early works~\citep{su2015multi,wei2020view} project unstructured point clouds into multiple 2D views, extract features from different views using 2D convolution, and then use sophisticated methods to fuse features from multiple views. MVCNN~\citep{su2015multi} is a pioneering work that uses a maximum pooling layer to aggregate multi-view information into global features. But the maximum pooling layer retains only the largest elements, which inevitably leads to information loss. To address this problem, Wei et al. proposed View-GCN~\citep{wei2020view} using directed graphs to find the relationship between individual views. Each individual view is regarded as a graph node, and the global shape descriptor is obtained by max pooling graph nodes of all levels.

Voxel-based approaches divide the point cloud into a uniform 3D spatial grid and use 3D convolutional neural networks for feature extraction. Zhou et al. introduced VoxelNet~\citep {zhou2018voxelnet} for robust 3D target detection. Although this method has achieved high detection performance, computation time and memory consumption sharply increase as the point cloud resolution is enhanced. To solve this problem, SECOND~\citep{yan2018second} proposed 3D sparse convolution that effectively reduces memory and computation costs, but most devices still have difficulty affording such a large amount of computation.

\subsection{Point-based Approaches}
PointNet~\citep{qi2017pointnet} has paved the way for relevant research studies on the point cloud, using MLP and max-pooling to extract and aggregate global features. PointNet++~\citep{qi2017pointnet++} introduces the concept of local features into 3D point cloud analysis. It uses Farthest Point Sampling (FPS) and ball query to perform centroid sampling, and neighborhood point search on point clouds to obtain various levels of local-global features. Subsequent work on 3D point cloud analysis focuses on the study of point cloud local feature aggregation descriptors. DGCNN~\citep{wang2019dynamic} builds a directed graph between centroids and neighboring points, extracts features of each edge using EdgeConv, and finally aggregates local features through max-pooling layer. In Geo-CNN~\citep{lan2019modeling}, the edge feature of each direction is weighted by a learnable matrix that relates to the direction. Then local features are aggregated according to the angle between the relative vector and three axes. 

Different from the above methods, the proposed GAM utilizes fine-grained geometric information to aggregate finer local features, which helps to improve the accuracy of subsequent tasks. Besides, compared to sophisticated local feature aggregation descriptors that lead to inefficiency in point cloud analysis models, the proposed GAM does not burden the computing device.

\section{Proposed Method}
The proposed gradient attention module (GAM) uses both gradient information and distance information between the center point and its neighboring points to generate corresponding importance weights of each neighboring point. The mathematical representation of gradient information of point cloud neighborhoods is given in Section 3.1, followed by the overall structure of GAM in Section 3.2. 

\subsection{Mathematical Representation of Point Cloud Neighborhood Gradients}\label{sec:math}
We use a center point and a point in its neighborhood as an example to illustrate the mathematical relation between gradient information, zenith angle, and azimuth angle. Given a set of N cloud points $\{{{p}_{i}}\} \in \mathbb{R}^{{N}\times 3}$,  where $i=1,2,... ,{N}$. A central point $p_i$ has the coordinate of $({{x}_{i}},{{y}_{i}},{{z}_{i}})$, and another point ${q}_j$ is the neighborhood point of ${p}_i$, whose coordinate is $({{x}_{j}},{{y}_{j}},{{z}_{j}})$. In the range image of the point cloud, two points $p_i$, $q_j$ can be represented as discrete points $f({{u}_{i}},{{v}_{i}})={{z}_{i}}$ and $f({{u}_{j}},{{v}_{j}})={{z}_{j}}$, and the conversion equation is written as follows.


\begin{equation}\label{eq:range}
\left\{\begin{aligned}
  &{{u}_{j}} =\frac{l}{d}{{x}_{j}}+{{u}_{0}}\\
  &{{v}_{j}} =\frac{l}{d}{{y}_{j}}+{{v}_{0}}
\end{aligned}\right.
\end{equation}

where $d$ is the depth of current point, $l$ is the camera focal length, ${{u}_{0}}$ and ${{v}_{0}}$ are the $X$, $Y$ coordinates of center point of the range image respectively.

Traditional method uses the difference of pixel value in depth map between current pixel and its adjacent pixels in the X and Y axis directions to represent depth gradient of the point. However, we focus on the association between the center point $p_i$ and one of its neighboring points $q_j$ in 3D space. Hence the point $q_j$ is regarded as a neighboring pixel of $p_i$ in the range image. We calculate the pixel value difference of these two points in the direction of an edge vector $\bm{\vec{b}}=({{u}_{j}}-{{u}_{i}},{{v}_{j}}-{{v}_{i}})$ to represent depth gradient of the point. The depth gradient $\nabla {{d}_{b}}$ is defined as $\nabla {{d}_{b}}=\frac{{{z}_{ji}}}{\sqrt{u_{ji}^{2}+v_{ji}^{2}}} $. And depth gradient along $X$, $Y$ axes $\nabla {{ d}_{x}}$, $\nabla {{d}_{y}}$ are defined as follows,

\begin{equation} \label{eq:gradient}
\left\{ \begin{aligned}
    \nabla {{d}_{x}} &=\frac{{{z}_{ji}}}{\sqrt{u_{ji}^{2}+v_{ji}^{2}}}*\frac{{{u}_{ji}}}{\sqrt{u_{ji}^{2}+v_{ji}^{2}}} \\ 
    \nabla {{d}_{y}}&=\frac{{{z}_{ji}}}{\sqrt{u_{ji}^{2}+v_{ji}^{2}}}*\frac{{{v}_{ji}}}{\sqrt{u_{ji}^{2}+v_{ji}^{2}}}. \\ 
\end{aligned} \right.
\end{equation}
where ${{u}_{ij}}$, ${{v}_{ij}}$, ${{z}_{ij}}$ denote ${{u}_{j}}-{{u}_{i}}$, ${{v}_{j}}-{{v}_{i}}$, ${{z}_{j}}-{{z}_{i}}$, respectively. Combining Eq.~\ref{eq:range} and Eq.~\ref{eq:gradient}, components of depth gradient along  $X$ and $Y$ axis are defined as$\nabla {{ d}_{x}}$, $\nabla {{d}_{y}}$.

\begin{equation}
\left\{ \begin{aligned}
 &\nabla {{d}_{x}}=\frac{d}{f}\frac{{{z}_{ji}}{{x}_{ji}}}{x_{ji}^{2}+y_{ji}^{2}}, \\ 
&\nabla {{d}_{y}}=\frac{d}{f}\frac{{{z}_{ji}}{{y}_{ji}}}{x_{ji}^{2}+y_{ji}^{2}}. \\    
\end{aligned} \right.  
\end{equation}

Since we explore the geometric structure of point clouds in 3D space, it is necessary to convert  depth gradient to the world coordinate system, in order to obtain gradients $\nabla  {{z}_{x}}$ and $\nabla {{z}_{y}}$ respectively. 

\begin{equation}
\left\{
\begin{aligned}{}
& \nabla {{z}_{x}}=\frac{d}{f}\nabla {{d}_{x}}=\frac{{{x}_{ji}}{{z}_{ji}}}{x_{ji}^{2}+y_{ji}^{2}},&\\
& \nabla {{z}_{y}}=\frac{d}{f}\nabla {{d}_{y}}=\frac{{{y}_{ji}}{{z}_{ji}}}{x_{ji}^{2}+y_{ji}^{2}}. \\
\end{aligned}\right.
\end{equation}

The approximate gradient $\bm{\vec{g}}$ of the neighboring point $q_j$ is $\bm{\vec{g}} =({{g}_{x}},{{g}_{y}},{{g}_{z}})$, where
\begin{equation}
\begin{aligned}
\left\{
\begin{array}{lll}
{g}_{x} & = \frac{{{z}_{ji}}}{\sqrt{x_{ji}^{2}+y_{ji}^{2}+z_{ji}^{2}}} \frac{{{x}_{ji}}}{\sqrt{x_{ji}^{2}+y_{ji}^{2}}},\\
{{g}_{y}} & = \frac{{{z}_{ji}}}{\sqrt{x_{ji}^{2}+y_{ji}^{2}+z_{ji}^{2}}}\frac{{{y}_{ji}}}{\sqrt{x_{ji}^{2}+y_{ji}^{2}}},\\
{{g}_{z}} & = \frac{\sqrt{x_{ji}^{2}+y_{ji}^{2}}}{\sqrt{x_{ji}^{2}+y_{ji}^{2}+z_{ji}^{2}}}.
\end{array}\right.
\end{aligned}
\end{equation}
Where $\frac{{{z}_{ji}}}{\sqrt{x_{ji}^{2}+y_{ji}^{2}+z_{ji}^{2}}}$ and $\frac{\sqrt{x_{ji}^{2}+y_{ji}^{2}}}{\sqrt{x_{ji}^{2}+y_{ji}^{2} +z_{ji}^{2}}}$ denote sine and cosine of the zenith angle of the neighborhood point respectively, $\frac{{{x}_{ji}}}{\sqrt{x_{ji}^{2}+y_{ji}^{2}}}$ and $\frac{{{y}_{ji}}}{\sqrt{x_{ji}^{2}+y_{ji}^{2}}}$ denote sine and cosine of the azimuth angle of the neighborhood point respectively. 

We simplify gradient calculation of neighborhood points by using the zenith angles and azimuth angles, and effectively reduce computation time.

\subsection{Gradient Attention Module} \label{sec:module}

In this section we introduce our proposed gradient attention module (GAM) based on the zenith and azimuth angles of neighborhood points, which includes gradient information of neighborhood points during neighborhood aggregation process. Therefore the model is enabled to capture more accurate local features, by using more fine-grained geometric information in the local feature aggregation descriptors. Details of our proposed GAM are given in Algorithm 1.

\begin{algorithm}[tb]
\caption{Gradient Attention Module}
\label{alg:algorithm}
\textbf{Input}: point cloud $\bm{P}=\{{\bm{p}_{i}}|i=1,... ,N \} \in {\mathbb{R}^{N\times 3}}$, with corresponding features $\bm{F}=\{{{\bm{f}_{i}}}|i=1,... ,N \} \in {\mathbb{R}^{N\times C}}$\\
\textbf{Parameter}: local feature extractor $\phi (\cdot )$, balanced weights $\lambda $, sampling radius r, number of centroid samples ${{N}_{s}}$, number of neighborhood point samples $K$\\
\textbf{Output}: output features $\bm{F}^{out}$
\begin{algorithmic}[1] 
\STATE {Sample ${{N}_{s}}$ points as the center points of the point cloud, with corresponding coordinates denoted as $\{{\bm{p}^{center}_{s}}|s=1,... ,{{{N}_{s}}}\} \in {\mathbb{R}^{{{N}_{s}}\times 3}}$.}
\STATE {Search $K$ points for each center point as its neighborhood, with corresponding coordinates $\bm{Q}^{NBR} =\{\bm{q}^{NBR}_{s,j}|s=1,... ,{{N}_{s}},j=1,... ,K \}\in {\mathbb{R}^{{{N}_{s}}\times K\times 3}}$, and corresponding features $\bm{F}^{NBR} = \{\bm{f}^{NBR}_{s,j}|s=1,... ,{{N}_{s}},j=1,... ,K \}\in {\mathbb{R}^{{{N}_{s}}\times K\times C}}$.}
\STATE {Create a directed graph in the neighborhood of each center point and compute relative position vector, distance information $\bm{d}_{s,j}$ (Eq.~\ref{eq:6}) and gradient information $\bm{g}_{s,j}$ (Eq.~\ref{eq:7}) of the neighborhood points. }
\STATE {Calculate weighted score matrix of neighborhood points $\bm{A}$ by using fine-grained geometric information $\bm{d}_{s,j}$ and $\bm{g}_{s,j}$.(Eq.~\ref{eq:8})}
\STATE {Local features $\bm{F}^{NBR}$ are fed to local feature extractor $\phi (\cdot )$, multiplied with the corresponding weight score matrix $\bm{A}$ and then weighted by $\lambda$ to obtain $\bm{F}^{out}$ (Eq.~\ref{eq:9}). }

\STATE \textbf{Return} $\bm{F}^{out}$
\end{algorithmic}
\end{algorithm}

As shown in Figure~\ref{fig: fig3}, there are a set of N points $ \bm{P}=\{{\bm{p}_{i}}|i=1,... ,N\}\in {\mathbb{R}^{N\times 3}} $ in the Cartesian coordinate system $(x,y,z)$, with their corresponding features  $\bm{F}=\{\bm{f}_{i}|i=1,... ,N \} \in {\mathbb{R}^{N\times C}}$. We use the baseline method~\citep{qi2017pointnet++, wang2019dynamic} to find center point and search for neighbor points. $\bm{P}^{center}=\{\bm{p}^{center}_{s}|s=1,... ,{{{N}_{s}}}\} \in {\mathbb{R}^{{{{N}_{s}}\times 3}}}$ represents the selected set of center point. For each center point, $K$ neighboring points are searched. In total there are $K*{N_s}$ neighboring points $\bm{Q}^{NBR}=\{  \bm{q}^{NBR}_{s,j}|s=1,... ,{{N}_{s}},j=1,... ,K \} \in {\mathbb{R}^{{{N}_{s}}\times K\times 3}}$ with corresponding features $\bm{F}^{NBR}=\{ \bm{f}^{NBR}_{s,j} |s=1,... ,{{{N}_{s}},j=1,... ,K\\}\}\in {\mathbb{R}^{{{N}_{s}}\times K\times C}}$, where $C$ denotes the number of channels of input features.

After establishing the directed graph between center points and each of its neighboring points, the relative position matrix is represent as $\bm{E} = \{ \bm{e}_{s,j} |s=1,... ,{{{N}_{s}},j=1,... ,K\\}\}$  .  $\bm{e}_{s,j}$ is represented as $\bm{q}^{NBR}_{s,j}-\bm{p}^{center}_{s}$, where $\bm{p}^{center}_s$ is the center point and $\bm{q}^{NBR}_{s,j}$ is one of its neighboring points. Vector length $\bm{{d}_{s,j}}$ is also expressed as follows,

\begin{equation} \label{eq:6}
\begin{aligned}
\left\{
\begin{array}{ll}
{\bm{e}_{s,j}}& =(\overrightarrow{\bm{x}_{s,j}},\overrightarrow{\bm{y}_{s,j}},\overrightarrow{\bm{z}_{s,j}}) \\ 
{\bm{d}_{s,j}}& =\sqrt{{{(\overrightarrow{\bm{x}_{s,j}})}^{2}}+{{(\overrightarrow{\bm{y}_{s,j}})}^{2}}+{{(\overrightarrow{\bm{z}_{s,j}})}^{2}}} \\
\end{array}\right.
\end{aligned}
\end{equation}
where $(\overrightarrow{\bm{x}_{s,j}},\overrightarrow{\bm{y}_{s,j}},\overrightarrow{\bm{z}_{s,j}})$ represent the relative position vector in the Cartesian coordinate system. 

Then the sum of azimuthal sine and cosine values of each neighboring point is calculated, and multiplied with sine of zenith angle, to represent the gradient information of the neighborhood points $\bm{{g}_{s,j}}$ written as follows,
\begin{equation} \label{eq:7}
\bm{g}_{s,j}=(\frac{\overrightarrow{\bm{z}_{s,j}}}{\bm{d}_{s,j}}\frac{\overrightarrow{\bm{x}_{s,j}}+\overrightarrow{\bm{y}_{s,j}}}{\sqrt{(\overrightarrow{\bm{x}_{s,j}})^{2}+(\overrightarrow{\bm{y}_{s,j}})^{2}}}).
\end{equation}
Our proposed GAM uses MLP to fuse gradient information and distance information of neighborhood points to obtain attentive weight calculated as follows,
\begin{equation} \label{eq:8}
a_{s,j}=Sigmoid(MLP([\bm{g}_{s,j};\bm{d}_{s,j}])).
\end{equation}
where Sigmoid is the activation function and [;] denotes the concatenation operation. The weight matrix is represented as $\bm{A} = \{ a_{s,j} |s=1,... ,{{{N}_{s}},j=1,... ,K\\}\}$

After obtaining the weight matrix $\bm{A}$, it is multiplied with the input point cloud features. Thus GAM can extract  deep features of the point cloud according to importance of each neighborhood point. A complete local feature aggregation process can be expressed as follows, 
\begin{equation} \label{eq:9}
\bm{F}^{out}=MLP(\frac{\lambda \phi ({\bm{F}^{NBR})\cdot {\bm{A}}+\phi ({\bm{F}^{NBR}}})}{1+\lambda})
\end{equation}
where $ \bm{F}^{out} \in {\mathbb{R}^{{{N}_{s}}\times K\times C_{out}}}$, $C_{out}$ denotes the number of channels of output features.  $\lambda$ is the balance weight and $\cdot$ represents the element-wise multiplication. $\phi (\cdot )$ denotes the local feature extractor used in previous works~\citep{ma2022rethinking, qi2017pointnet++, wang2019dynamic} to extract deeply aggregated features.



\section{Experiments}
To fully evaluate the effectiveness and generality of our proposed GAM, we apply our proposed method on several state-of-the-art methods for 3D shape classification, 3D part segmentation, 3D semantic segmentation, and 3D object detection. Experiments are conducted on S3DIS dataset~\citep{armeni20163d}, ScanObjectNN dataset~\citep{uy2019revisiting}, ShapeNet dataset~\citep{yi2016scalable}, KITTI dataset~\citep{geiger2013vision} and ModelNet40 dataset~\citep{wu20153d}, respectively. 
The same training strategy used in each baseline is employed in our experiments, except that only GAM is added in each downsampling layer of the model. And $\lambda$ is set to 1. The number of channels of the two-layer MLP in GAM is set to (1,16), (16,1). Experiments are run on NVIDIA GTX 3090 GPU and AMD EPYC 7402 CPU.

\begin{table}[]
\small
\setlength\tabcolsep{3pt}
\centering

\begin{tabular}{c|ccc|ccc}
\hline
\makebox[0.08\textwidth][c]{Method} & \makebox[0.04\textwidth][c]{mIoU} & \makebox[0.04\textwidth][c]{OA} & \makebox[0.04\textwidth][c]{mAcc} & \makebox[0.033\textwidth][c]{TP} \\
\hline
PointNet~\cite{qi2017pointnet}            & 47.6      & 78.5    & 66.2      & -            \\
PointCNN~\cite{li2018pointcnn}            & 65.4      & 88.1    & 75.6      & -                  \\
PointWeb~\cite{zhao2019pointweb}          & 66.7      & 87.3    & 76.2      & -                        \\
RandLA-Net~\cite{hu2020randla}          & 70.0      & 88.0    & 82.0      & -                     \\
KPConv~\cite{thomas2019kpconv}        & 70.6      & -       & 79.1      & -                          \\
BAAF-Net~(Qiu et al. 2021)         & 72.2      & 88.9    & 83.1         & -          \\
\hline
PointNet++~\cite{qi2017pointnet++}        & 54.5      & 81.0    & 67.1      & 130                     \\
+ GAM                      & 56.6      & 81.8    & 71.7      & 127                      \\
\hline
DGCNN~\cite{wang2019dynamic}           & 56.1      & 84.1    & -         & 32                   \\
+ GAM                           & 58.8      & 85.5    & 69.1      & 31                        \\
\hline
Point Trans.~\cite{zhao2021point} & 73.5      & 90.2    & 81.9      & 26                       \\
+ GAM               & 73.9      & 89.9    & 83.0      & 25                        \\
 + GAM*              & \textbf{74.4}      & \textbf{90.6}    & \textbf{83.2}      & -          \\ 
\hline      
\end{tabular}
\caption{Comparison results on S3DIS dataset for 3D semantic segmentation with 6-fold cross-validation. Mean Inter-over-Union (mIoU), overall accuracy (OA) and mean accuracy (mAcc) are used as evaluation metrics. * represents the voting strategy and throughput using the test result in Area5. } 
\label{tab:tab1}
\end{table}

\subsection{Results on 3D Semantic Segmentation}
We conduct the 3D semantic segmentation experiment on S3DIS dataset~\citep{armeni20163d}, which contains 3D scanned point clouds of six interior regions with 272 rooms in total. Each point belongs to one of the 13 semantic categories containing wood panels, bookcases, chairs, ceilings, etc. We compare the proposed GAM with state-of-the-art methods, including PointNet~\citep{qi2017pointnet}, PointCNN~\citep{li2018pointcnn}, PointWeb~\citep{zhao2019pointweb}, RandLA-Net~\citep{hu2020randla}, KPConv~\citep{thomas2019kpconv}, BAAF-Net~\citep{qiu2021semantic}, PointNet++~\citep{qi2017pointnet++}, DGCNN~\citep{wang2019dynamic}, and Point Trans.~\citep{zhao2021point}. We select PointNet++, DGCNN, and Point Trans. as baselines to evaluate the effectiveness of adding our proposed method GAM.

\begin{figure*}[t]
    \centering
    \includegraphics[width=\linewidth]{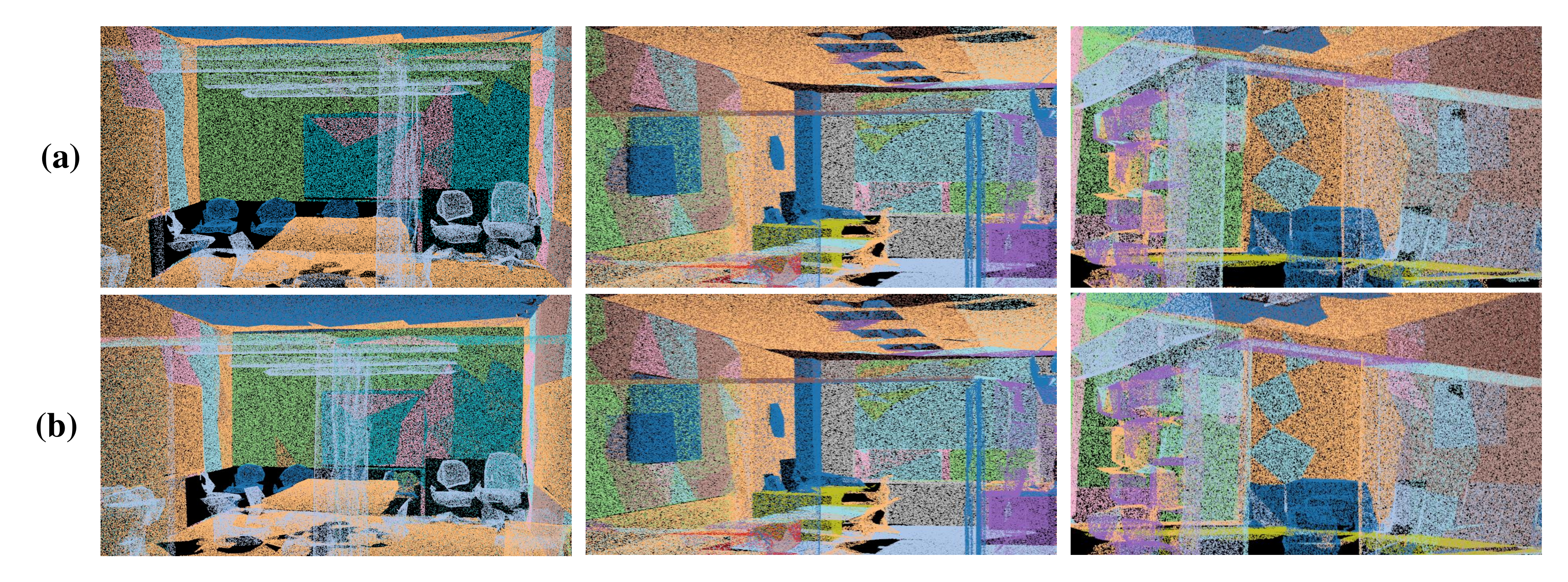}
    \caption{3D semantic segmentation experiment visualization results. (a) represents ground truth and (b) represents forecast results of point transformer~\cite{zhao2021point} after GAM is inserted.}
    \label{fig: fig4}
\end{figure*}

In Table~\ref{tab:tab1}, we report mean Inter-over-Union (mIoU), overall accuracy (OA) and mean accuracy (mAcc), and throughput(TP) for the 6-fold cross-validation of S3DIS dataset. We can find that after adding GAM into the three baselines PointNet++, DGCNN, and Point Trans, mIoU scores increase 2.1\%, 2.7\%, and 0.4\% respectively, OA scores and mAcc scores are also significantly increased. After using Voting, GAM with Point Trans achieves the best performance, where mIoU score, OA score, and mAcc score are 74.4\%, 90.6\%, and 83.2\%, respectively. Inference time of three baselines after adding GAM only increases by 2.3\%, 3.1\%, and 3.8\%, respectively. Increment on computational costs in terms of throughput is almost negligible. These experimental results demonstrate that our proposed GAM is lightweight and efficient. Figure~\ref{fig: fig4} shows visualization result of 3D semantic segmentation of S3DIS dataset.

\begin{table}[]
\small
\setlength\tabcolsep{3pt}
\centering

\begin{tabular}{c|cc|ccc}
\hline
\makebox[0.08\textwidth][c]{Method} & \makebox[0.05\textwidth][c]{OA} & \makebox[0.05\textwidth][c]{mAcc} & \makebox[0.04\textwidth][c]{TP} \\
\hline
PointNet++~\cite{qi2017pointnet++}    & 77.9      & 75.4      & -                          \\
DGCNN~\cite{wang2019dynamic}      & 78.1      & 73.6      & -                             \\
GBNet~\cite{qiu2022geometric}   & 80.5      & 77.8      & -                            \\
\hline
PRA-Net(1K)~\cite{cheng2021net} & 81.0      & 77.9      & 1152                       \\
+ GAM                & 81.6      & 77.9      & 1094                        \\
\hline
Point-TNT~(Berg et al. 2022)  & 83.6 & 82.3 & 240                      \\
+ GAM                  & 85.0 & 82.8 & 238                          \\
\hline
PointMLP-elite~\cite{ma2022rethinking} & 84.4 & 82.6 & 749                     \\
+ GAM             & 84.8 & 88.24 & 708                         \\
\hline
PointMLP~\cite{ma2022rethinking}    & 85.7 & 84.4 & 213                       \\
+ GAM                   & 86.1 & 84.7 & 204                        \\
\hline
RepSurf-2x~\cite{ran2022surface}  & 86.0      & -         & 420                         \\
+ GAM                 & 86.4      & -         & 418                          \\
\hline
PointNeXt-s~\cite{qian2022pointnext} & 88.1 & 86.4 & 1628                       \\
+ GAM                & \textbf{88.4} & \textbf{86.5} & 1544                 \\
\hline
\end{tabular}
\caption{Comparison result on ScanObjectNN dataset for 3D shape classification. For a fair comparison, all methods in the table use 1024 points as the input. Overall accuracy (OA) and mean accuracy (mAcc) are used as evaluation metrics. }

\label{tab:tab2}
\end{table}

\subsection{3D Shape Classification Experimental Results on ScanObjectNN}
We conduct the 3D shape classification experiment on  ScanObjectNN dataset~\citep{uy2019revisiting} that contains 15,000 objects of 15 different classes. The most difficult and commonly used variant of ScanObjectNN $PB_T50_RS$ is implemented for experiments. We compare our proposed GAM with state-of-the-art methods, including PointNet++~\citep{qi2017pointnet++}, DGCNN~\citep{wang2019dynamic}, GBNet~\citep{qiu2022geometric}, PRA-Net(1k)~\citep{cheng2021net}, Point-TNT~\citep{berg2022points}, PointMLP-elite~\citep{ma2022rethinking}, PointMLP~\citep{ma2022rethinking}, RepSurf-2x~\citep{ran2022surface} and PointNeXt-s~\citep{qian2022pointnext}. Among these methods, PRA-Net(1k), Point-TNT, PointMLP-elite, PointMLP, RepSurf-2x, and PointNeXt-s are used as baselines to evaluate the effectiveness of adding our proposed GAM. 

In Table~\ref{tab:tab2}, we report the overall accuracy (OA) score, mean accuracy (mAcc), and throughput (TP) on the ScanObjectNN dataset. The voting strategy is not used in each baseline. After adding GAM, the OA score and mAcc score of each baseline method are increased significantly. Experiment results demonstrate that our proposed GAM has the potential to be widely used in the point cloud domain. Besides, results in terms of throughput (TP) indicate that computation cost barely increases after adding GAM.

\begin{table}[]
\centering
\begin{tabular}{c|ccc}
\hline
\makebox[0.06\textwidth][c]{Method} & \makebox[0.04\textwidth][c]{Cars} & \makebox[0.06\textwidth][c]{Pedestrians} & \makebox[0.06\textwidth][c]{Cyclists} \\
\hline
IA-SSD~\cite{zhang2022not} & \textbf{79.57}    & 58.91           & 71.24        \\
+GAM    & 79.16    & \textbf{59.31}           & \textbf{72.58}        \\
\hline
\end{tabular}
\caption{Comparison results of 3D object detection on KITTI 3D target detection validation set. Mean Inter-over-Union (mIoU) score is used as the evaluation metric. }

\label{tab:tab3}
\end{table}

\subsection{Result on 3D Object Detection} 
For 3D object detection task, experiments are conducted on the KITTI dataset~\citep{geiger2013vision}, which has three detection categories, cars, pedestrians, and bicycles. Each category has three subsets, "easy", "medium" and "difficult", basing on the detection difficulty. The "medium" subset is the most commonly used for evaluation. 

In Table~\ref{tab:tab3}, we report mean Inter-over-Union (mIoU) score on validation set of the KITTI dataset. After adding GAM, mIoU scores of pedestrian and bicycle category are improved by 0.4\% and 1.34\% respectively, while for car category mIoU score decreases by 0.41\%. These results indicate that detection performance for small targets can be effectively improved with GAM.

\subsection{3D Shape Classification Experimental Results on ModelNet40}
\begin{table}[]
\begin{tabular}{c|c|cc}
\hline
\makebox[0.02\textwidth][c]{Method} & \makebox[0.022\textwidth][c]{Input} & \makebox[0.025\textwidth][c]{OA} & \makebox[0.025\textwidth][c]{mAcc} \\
\hline
KPConv~\cite{thomas2019kpconv}      & 7k   & 92.9      & -         \\
Point Trans.~\cite{zhao2021point}             & 1k    & 93.7      & 90.6      \\
CurveNet~\cite{xiang2021walk}                 & 1k    & 94.2      & -         \\
\hline
PointNet++(S)~\cite{qi2017pointnet++}  & 1k   & 92.2      & 89.1      \\
+ GAM                              & 1k   & 92.8      & 91.5      \\
\hline
PointNet++(M)~\cite{qi2017pointnet++} & 1k+N & 92.8      & 90.7      \\
+ GAM & 1k+N & 93.3      & 91.4      \\
\hline
DGCNN~\cite{wang2019dynamic}                        & 1k   & 92.9      & 90.2      \\
+ GAM & 1k   & 93.3      & 90.5      \\
\hline
PointMLP*~\cite{ma2022rethinking}                    & 1k   & 94.5      & 91.4      \\
+ GAM*& 1k    & \textbf{94.7}      & \textbf{91.9} \\
\hline
\end{tabular}
\caption{Comparison results on ModelNet40 dataset on 3D shape classification tasks. N indicates that the input point cloud contains normal vector information, * indicates that voting is used, S represents Single-Scale Grouping, and M represents Multi-Scale Grouping. }
\label{tab:tab4}
\end{table}

We conduct 3D shape classification experiments on ModelNet40 dataset~\citep{wu20153d}, which has 12311 CAD samples, including 9843 training samples and 2468 test samples. The proposed GAM is compared with the state-of-the-art methods, which are KPConv~\citep{thomas2019kpconv}, Point Transformer~\citep{zhao2021point}, CurveNet~\citep{xiang2021walk}, PointNet++(SSG)~\citep{qi2017pointnet++}, PointNet++(MSG)~\citep{qi2017pointnet++}, DGCNN~\citep{wang2019dynamic}, and PointMLP~\citep{ma2022rethinking}. We select PointNet++(SSG), PointNet++(MSG), DGCNN, and PointMLP as baselines to evaluate effectiveness of our proposed method. 

In Table~\ref{tab:tab4}, we report overall accuracy (OA) score and mean accuracy (mAcc) on the ModelNet40 dataset. After adding GAM, OA score and mAcc are increase significantly.

\begin{table}[]
\setlength\tabcolsep{4pt}
\centering
\begin{tabular}{c|cc}
\hline
\makebox[0.28\textwidth][c]{Method} & \makebox[0.07\textwidth][c]{Ins} & \makebox[0.07\textwidth][c]{TP}\\
\hline
PointNet~\citep{qi2017pointnet}     &83.7     &-\\
Point Tran.~\citep{zhao2021point}  &86.6  &-\\
PointMLP~\citep{ma2022rethinking}    &86.1  &-\\
\hline
PointNet++~\citep{qi2017pointnet++}  &85.1  &370\\
+ GAM       &85.5  &368\\
\hline
DGCNN~\citep{wang2019dynamic}       &85.2  &257\\
+ GAM       &85.5  &235\\
\hline
CurveNet~\citep{xiang2021walk}*   &86.8  &104\\
+ GAM*      &\textbf{87.0}  &99\\   
\hline
\end{tabular}
\caption{Comparison results on the ShapeNet dataset for 3D part segmentation on different classes. Ins. represent the instance average Inter-over-Union. * denotes the use of voting.}
\label{tab:tab5}
\end{table}

\subsection{Result on 3D Part Segmentation}


We conduct 3D part segmentation experiments on the ShapeNet dataset~\citep{yi2016scalable} that has 16881 3D objects with 50 different categories of segmentation masks.  We compare our GAM with the state-of-the-art methods, including  PointNet~\citep{qi2017pointnet}, Point Tran.~\citep{zhao2021point}, PointMLP~\citep{ma2022rethinking}, PointNet++~\citep{qi2017pointnet++}, DGCNN~\citep{wang2019dynamic} and CurveNet~\citep{xiang2021walk}. Each object was sampled to 2048 points. 

In Table~\ref{tab:tab5}, we report the instruction mIou (Ins.) and throughput (TP) for each baseline method before and after adding GAM. The Ins. scores of baseline methods (i.e. PointNet++, DGCNN, and SOTA CurveNet) increase by 0.4\%, 0.3\%, and 0.2\% after adding GAM, respectively. 



\subsection{Ablation Study}

In order to verify the effectiveness of gradient information, two variants of GAM are created, one with distance information and the other with gradient information. Results of ablation experiments conducted on the ScanObjectNN dataset and S3DIS dataset are present in Table~\ref{tab:tab6} and Table~\ref{tab:tab7}, respectively.
\begin{table}[]
\setlength\tabcolsep{4pt}
\centering
\begin{tabular}{cc|cc|cc}
\hline
\makebox[0.05\textwidth][c]{ } & \makebox[0.05\textwidth][c]{  } & \multicolumn{2}{c|}{PointNet++} & \multicolumn{2}{c}{PointMLP} \\
\hline
Distance & Gradient & OA      & mA       & OA        & mA     \\
\hline
\XSolid         & \XSolid         & 86.2          & 84.3           & 85.7     & 84.4    \\
\Checkmark        & \XSolid         & 86.1          & 84.2           & 84.3     & 82.4    \\
\XSolid         & \Checkmark        & 86.5          & 84.5           & 85.8     & 84.6    \\
\Checkmark        & \Checkmark        &  \textbf{87.0}          &  \textbf{85.8}           &  \textbf{86.1}     &  \textbf{84.7}    \\
\hline
\end{tabular}
\caption{Ablation study of our proposed GAM using different kinds of information on ScanObjectNN dataset.}
\label{tab:tab6}
\end{table}

\begin{table}[]
\centering
\begin{tabular}{cc|cc|cc}
\hline
         &          & \multicolumn{2}{c|}{PointNet++} & \multicolumn{2}{c}{DGCNN} \\
\hline
Distance & Gradient & mIoU        & OA       & mIoU        & OA     \\
\hline
\XSolid        & \XSolid        & 53.5          & 83.0           & 47.9     & 83.6    \\

\Checkmark        & \XSolid        & 54.1         &  \textbf{83.2}           & 49.3     & 84.4    \\
\XSolid        & \Checkmark        & 54.5          & 83.0           & 49.7     & 84.5    \\
\Checkmark        & \Checkmark        &  \textbf{54.8}         &  \textbf{83.2}           &  \textbf{50.0}     &  \textbf{84.6} \\
\hline
\end{tabular}
\caption{Ablation study of our proposed GAM using different kinds of information on S3DIS dataset Area5.}
\label{tab:tab7}
\end{table}



As shown in Table~\ref{tab:tab6}, after adding the GAM variant with only distance information for PointNet++ and PointMLP, OA scores decrease by 0.1\% and 1.3\%. After adding the GAM variant with only gradient information, the OA score increases by 0.3\% and 0.1\%, indicating that gradient information can constrain the local feature aggregation process more effectively. While with both distance and gradient information, the OA score can be significantly increased than single fine-grained information. Therefore, we conclude that the combination of various fine-grained geometric information is more effective in constraining the local aggregation process by using multiple geometric dimensions.

In Table~\ref{tab:tab7}, we report the results of the ablation study for our proposed GAM using different kinds of information on the S3DIS dataset Area5. It also demonstrates the superiority of gradient information over distance information, and the effectiveness of combining various fine-grained geometric information.

\begin{table}[ht]
\centering
\begin{tabular}{c|c|c}
\hline
       & Normal & Zenith \&   Azimuth \\
\hline
time (ms) & 18.6   & 0.522 \\
\hline
\end{tabular}
\caption{Computation time of a single run before and after GAM simplification. Normal represents direct calculation of the normal vector for each neighborhood point, Zenith \& Azimuth represents calculation of the explicit representation using the zenith angles and azimuth angles.}
\label{tab:tab8}
\end{table}


In three-dimensional space, the gradient of a three-dimensional function and the normal vector of a three-dimensional isosurface have different geometric meanings, but they are essentially the same. Therefore, we modify the local surface fitting method~\citep{pauly2003point} to calculate the normal vector of a plane, which is formed by the projection of the center point on the X and Y circles and neighboring points. It is the same as the object meaning of the gradient calculated in GAM. The results are shown in Table~\ref{tab:tab8}. After simplification, calculation speed is about 35 times faster than the original method, which effectively alleviates the problem of slow inference speed after adding GAM.


\section{Conclusion}
In this paper, we propose an efficient, lightweight, and plug-and-play gradient attention module (GAM), in which gradient information is introduced into the local feature aggregator for the first time. Our proposed GAM solves the problem of the different importance of each neighborhood point in the local feature aggregation process, and brings fine-grained geometric information to the local aggregation process. The effective and efficient performance of our proposed GAM is verified by conducting comparison experiments on four tasks, including 3D point cloud shape classification, 3D part segmentation, 3D semantic segmentation, and 3D object detection. It is our expectation that this work can promote further research on local feature aggregation descriptors.

\section{Acknowledgements}
This work was supported by a grant from Zhejiang Leapmotor Technology CO., LTD, China.
\bibliography{aaai23}
\end{document}